\documentclass[10pt,twocolumn,letterpaper]{article}
\pdfoutput=1
\usepackage{cvpr}
\usepackage{times}
\usepackage{epsfig}
\usepackage{graphicx}
\usepackage{amsmath}
\usepackage{amssymb}
\usepackage[ruled,linesnumbered]{algorithm2e}
\usepackage{mathtools}
\usepackage{booktabs}
\usepackage{color}
\usepackage{tabu}
\usepackage[nocompress]{cite}

\usepackage[breaklinks=true,bookmarks=false]{}

\cvprfinalcopy 

\def\cvprPaperID{2} 

\newcommand{\MOTNEW}{{\it MOT16}\xspace}
\newcommand{\MOTLATEST}{{\it MOT17}\xspace}

\DeclareMathOperator*{\argmin}{arg\min}

\DeclareMathOperator{\FW}{FW}
\DeclareMathOperator{\round}{BINARIZE}

\DeclareMathOperator{\IMAX}{IMAX}

\newcommand{\Fig}{Fig.\xspace}
\newcommand{\norm}[1]{\left\lVert#1\right\rVert_{2}}
\newcommand{\Tab}{Tab.\xspace}
\DeclareMathOperator{\logit}{logit}

\newcommand{\vect}[1]{\mathbf{#1}}
\newcommand{\mat}[1]{\mathbf{#1}}
\newcommand{\graph}{\mathcal{G}} 
\newcommand{\numPersons}{P} 
\newcommand{\vertexSet}{\mathcal{V}} 
\newcommand{\edgeSet}{\mathcal{E}} 
\newcommand{\pairwCosts}{q_{u,v}} 
\newcommand{\unaryCosts}{c_{\vertex}} 
\newcommand{\arbLabel}{k} 
\newcommand{\decisionVar}{x} 

\newcommand{\iterateVecBinarized}{\vect{\decisionVar}_{b}}
\newcommand{\iterateVecBinarizedIndexed}[1]{\iterateVecBinarized(#1)}
\newcommand{\iterateVecIndexed}[1]{\vect{\decisionVar}(#1)}
\newcommand{\atomicVec}{\vect{a}}
\newcommand{\diffVec}{\vect{d}(j)}
\newcommand{\atomicVecIndexed}[1]{\atomicVec(#1)}
\newcommand{\vertex}{v} 
\newcommand{\vertu}{u} 
\newcommand{\qMat}{\mat{Q}} 
\newcommand{\modelFeatures}{\mathcal{F}} 
\newcommand{\regulizer}{r} 
\newcommand{\neighborhood}{\mathcal{N}} 
\newcommand{\hierarchicalIterate}[1]{\vect{\decisionVar}^{(#1)}} 
\newcommand{\arbdecisionVar}{\decisionVar_{\vertex}^{\arbLabel}} 

\graphicspath{{figures/}}

\definecolor{darkgreen}{rgb}{0,.75,0}

\begin{document}

\title{Fusion of Head and Full-Body Detectors for Multi-Object Tracking}

\author{
Roberto Henschel$^1$ \quad
Laura  Leal-Taix\'{e}$^2$ \quad
Daniel Cremers$^2$ \quad
Bodo Rosenhahn$^1$ \\ \vspace{0.2cm}
\small{$^1$Leibniz Universit\"at Hannover \quad$^2$Technische Universit\"at M\"unchen } 
\\ \small{\{henschel,rosenhahn\}@tnt.uni-hannover.de \quad \{leal.taixe,cremers\}@tum.de}}

\maketitle

\begin{abstract}
   In order to track all persons in a scene, the tracking-by-detection paradigm has proven to be a very effective approach.  Yet, relying solely on a single detector is also a major limitation, as useful image information might be ignored.
    Consequently, this work demonstrates how to fuse two detectors into a tracking system. 
   To obtain the trajectories, we propose to formulate tracking as a weighted graph labeling problem, resulting in a binary quadratic program. As such problems are NP-hard, the solution can only be approximated. Based on the Frank-Wolfe algorithm, we present a new solver that is crucial to handle such difficult problems.
   Evaluation on pedestrian tracking is provided for multiple scenarios, showing superior results over single detector tracking and standard QP-solvers. Finally, our tracker ranks $2$nd on the MOT16 benchmark and $1$st on the new MOT17 benchmark, outperforming over 90 trackers.
\end{abstract}


\section{Introduction}
\label{sec:introduction}

Multiple object tracking, and in particular people tracking, is one of the key problems in computer vision with potential impact for many applications such as video surveillance or crowd analysis \cite{alahi2014socially}. 
A common approach to generate the trajectories of multiple people is
\emph{tracking-by-detection}: first a person detector is applied to
each individual frame to find the putative locations of people. Then,
these hypotheses are linked across frames to form trajectories.
By building on the advances in person detection over the last decade,
tracking-by-detection has been very successful \cite{choiiccv2015,dehghancvpr2015,lealcvpr2014,tangBMTT2016}. 
However, the dependence on detection results, typically bounding boxes, is also a major limitation.
A lot of potentially useful information is lost during the non-maxima suppression. A tracker typically does not use direct image data, except in the form of appearance models in order to discriminate different people.
Recently, a number of approaches  \cite{chencvpr2014,fragkiadakieccv2012,milancvpr2015} have proposed to use other image features aside from full-body detections, with the main goal of recovering partially occluded pedestrians.

\begin{figure}[t]
\centering  
\def\svgwidth{1.05\linewidth}
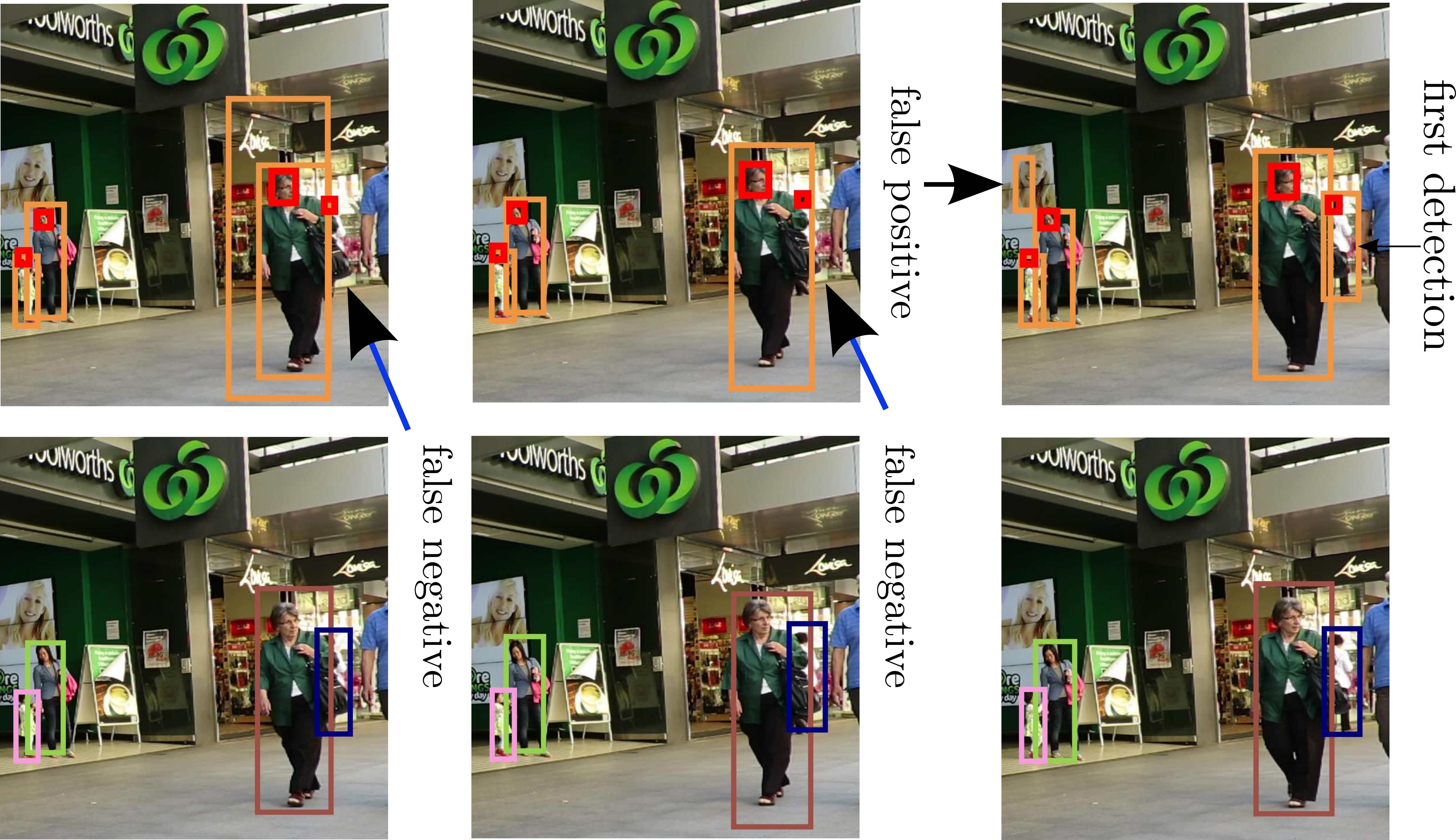
\caption{MOT16-09 sequence (from left to right: frame 10,12,15). \textbf{Top row}: Body detections (orange) and head detections (red). The body detector misses the person depicted by the arrow until frame 15. \textbf{Bottom row}: The result of our tracker. The false positive is removed as it does not have a corresponding head detection. The tracker recovers the heavily occluded pedestrian due to the presence of head detections. 
   } 
\label{fig:teaserReal}  
\end{figure}

In this paper, we present a framework for offline multiple object tracking
using two detector types, namely,  full-body detections together with head detections, since heads can be detected very accurately, as they are barely prone to pose variations or occlusions. 
This is especially useful in crowded scenarios: Fig.~\ref{fig:teaserReal} shows a heavily occluded pedestrian. While the full-body detector is unable to detect that person, due to the occlusion, its head is still visible so that our tracker localizes that pedestrian correctly.

Our tracking formulation ensures long-term temporal consistency by taking \textit{all} detections assigned to a person (we denote detection to person assignments as labelings) into account. 
Therefore, our clustering concept shares similarities to correlation clustering approaches  \cite{tang2015subgraph,tangBMTT2016,zamireccv2012,dehghancvpr2015}, but we propose a very efficient labeling formulation that avoids the exponential growth in the constraints. 
Due to our powerful solver, we are able to optimize our problem globally on the input detections without the need of potentially error-prone tracklets.

We compute the best labeling by solving a Binary Quadratic Problem (BQP). A straightforward approach to solve that BQP would thus be to optimize an equivalent Binary Linear Program (BLP) using branch\&bound. However, due to the high dimensionality of the problem, such a BLP is computationally expensive and memory demanding. 

We propose to use the Frank-Wolfe algorithm (FW) to solve the relaxation of the BQP. 
By using a standard implementation of FW, the result is often far away from the binary optimal solution.
Therefore, we propose several crucial improvements that lead in practice to a much better solution in terms of the objective value and the tracking performance, as we show in Section~\ref{FWexperiments}.
At the same time, the proposed
 algorithm is much faster than the standard 
branch\&bound approach.
Finally, an analysis on the effect of the fusion of head detections with full-body detections shows that the best tracking accuracy is obtained by using both input sources. The fusion helps especially to remove false positive full-body detections that are not consistent with the head detections and to recover heavily occluded persons. 

\subsection{Contributions}

To summarize, our contribution is three-fold:

\begin{itemize}
   \item We propose a novel detector fusion multi-object tracking system, which solves a graph labeling problem and is represented by a BQP with very few constraints. 
	\item We propose a new solver that significantly  improves  over standard BQP solvers when applied to our discrete optimization problem.
   \item  We present detailed evaluations on the improvements  due to our solver as well as the detector fusion. Our framework sets a new state-of-the-art in tracking.
\end{itemize}

\subsection{Related Work}
\label{sec:related_work}

\noindent{\bf Data association models.} 
Tracking-by-detection has become the standard paradigm for multi-object tracking. It splits the problem into two steps:
 object detection and
data association. In crowded environments, where occlusions are
common, even state-of-the-art detectors
\cite{dollartpami2014,felzenszwalb2010object,rennips2015,yang2016exploit} are prone to false
alarms and missed detections. 
The goal of the data association step is then to fill in the gaps between detections and filter out false positives.
In order to do this robustly, data association is mostly performed for all frames and all trajectories simultaneously.
This is usually done in discrete space, using graph based methods \cite{HenLea2014a,solera2015learning,zhangcvpr2008,berclaztpami2011,pirsiavashcvpr2011,brendelcvpr2011},
or BLPs \cite{jiangcvpr2007,charicvpr2015}.

Most of these trackers were derived from a Markov chain model \cite{zhangcvpr2008}.  Recent systems utilize correlation clustering based formulations that ensure consistency within all links of a trajectory \cite{dehghancvpr2015,milancvpr2015,tang2015subgraph,tangBMTT2016,tang2017multiple,tesfaye2016multi,zamireccv2012}.  Thereby, simplified models were used initially, which created trajectories iteratively, computing one best clique \cite{zamireccv2012} or dominant set \cite{tesfaye2016multi} corresponding to exactly one person and then removing the respective detections from the loop. This concept has been extended \cite{dehghancvpr2015,milancvpr2015} to obtain trajectories in a global manner, for all persons at the same time. However, the inference relies on potentially error-prone initial tracklets to keep the approach computationally feasible.  %
In contrast, our solver is fast and accurate enough to optimize directly on the detections, thereby avoiding error propagation that might have been introduced by the tracklets.
Further progress has been made by computing the correlation clustering directly on the input detections \cite{tang2015subgraph,tangBMTT2016}, using a huge set of clique constraints in a BLP, that has exponential growth. Accordingly, a heuristic solver has to be applied.  
In contrast, our formulation needs only very few constraints, making it capable for the usage of many detections.  

\noindent{\bf BQP Optimization.} Tracking methods that need to solve a BQP have been rare so far, due to the computational challenge, although many advanced tracking models are naturally expressed as a BQP. 
For instance, the Markov model \cite{zhangcvpr2008} can be augmented by one additional detector \cite{charicvpr2015}, resulting in a BQP.   While this problem can be solved by rewriting the BQP as an equivalent BLP, we show in our experiments, that this simple trick is not applicable to our more demanding correlation clustering based model, due to the problem size of our BQP.  
Another work  \cite{dehghan_binary_2016} formulates online tracking via a BQP and solves it using the Frank-Wolfe algorithm, which is also the basis for our solver.
While \cite{dehghan_binary_2016} shows good performance, we propose a hierarchical solving scheme that can be easily integrated into their formulation, thereby further improving their result.  Furthermore, during the Frank-Wolfe algorithm, the step size for an iterate update has to be computed. We derive an optimal, algebraic computation, that is cheap to compute and improves over existing methods \cite{assari2016re,dehghan_binary_2016,DBLP:journals/corr/abs-1207-4747}. Note that our improvements may be applied to methods of other fields in computer vision as well, such as person re-identification \cite{assari2016re}, co-localization \cite{joulin2014efficient} or object segmentation \cite{seguin_instance-level_2016}. %

\noindent{\bf Incorporating different features.} 
Limiting the input of the tracker to a single detector has clearly several drawbacks, since much of the information of the image is not taken into account, potentially ignoring semi-occluded objects. 
In recent literature, several works have started incorporating different image features for the task of multi-target tracking.
Few works use supervoxels as input for tracking, obtaining as a byproduct a silhouette of the pedestrian.  
In \cite{chencvpr2014}, the optimization is done via greedy propagation, while in \cite{milancvpr2015}, supervoxel labeling is formulated as CRF.

There are several works that use dense point tracks (DPT) \cite{broxeccv2010} or KLT  \cite{tomasi1991detection,lucas1981iterative}  together with detections to improve tracking performance. In \cite{benfoldcvpr2011}, corner features are tracked using KLT to obtain a motion model between detections. 
In \cite{fragkiadakicvpr2011}, multi-target
tracking is tackled by clustering DPTs and further combined with detection-based tracklets in a two-step approach in \cite{fragkiadakieccv2012}. Further improvement is achieved using a globally optimal fusion formulation \cite{henschel2016tracking}.
In \cite{charicvpr2015}, a BQP fuses head and body detections to track pedestrians, modeling non-maxima suppression as well as overlap consistency between features. 
In contrast to our model, only co-occurrences of active features are considered, while we directly model the grouping of features to different persons, allowing to ensure consistency within each cluster over long time periods.
Also in the extension \cite{seguin_instance-level_2016} to motion segmentation using superpixels, the per-person consistency is not considered. 
%


\vspace{-0.1cm}
\section{Detector Fusion for Multi-Target Tracking}
\label{sec:optimization}

 In this section, we describe the data association that couples multiple detectors and detections in a correlation clustering fashion to ensure long-term temporal consistency. 
As correlation clustering is NP-Complete \cite{bansal2004correlation}, we rely on finding a good approximation to the solution. We propose to use a BQP formulation for the clustering problem that can be well approximated using the Frank-Wolfe \cite{frankwolfe} solver. In particular, we compute the relaxed solution of the BQP first, and perform a rounding step afterwards. 
Frank-Wolfe is well suited for continuous quadratic
problems with linear constraints, as each iteration step involves solving
a computationally efficient linear optimization problem. %
The binary solution is then obtained by an efficient rounding step. 

When applied to a non-convex problem, like our model, the Frank-Wolfe algorithm delivers only a local optimum \cite{lacoste2016convergence}. 
Hence, simply applying the standard algorithm will result in a solution that is far away from the global optimum.  %
We thus focus on enhancing the solution of Frank-Wolfe by: (i) regularizing the cost function, (ii)
 computing the optimal step size within the solver's algorithm algebraically and (iii) introducing a hierarchical solving scheme that enhances the solution produced by the Frank-Wolfe algorithm. 

Our regularizer prevents the Frank-Wolfe algorithm from falling to quickly into a local optimum.
The hierarchical solving scheme gains the improvement by revoking or connecting clusters of the discretized solution, while having the guarantee of operating optimally. 
The presented approach is not specific to the Frank-Wolfe solver. It can be applied after any  approximating algorithm. It further allows to correct errors introduced by the initial solver.

Experiments in Sect.~\ref{sec:experiments} show that our proposed solver provides good solutions close to the estimated bound, while being considerably faster than  the commercial solver Gurobi \cite{Gurobi},  which uses the branch-and-bound/cut algorithm \cite{land1960automatic,mitchell2002branch} to find the globally optimal solution.

\subsection{Joint Data Association}
\label{subsec:JDA}
We cast the data association using two detectors as a graph labeling problem: Consider a weighted complete graph $\graph=(\vertexSet,\edgeSet,c)$, where the vertex set $\vertexSet$ consists of all input detections. We set $n=|\vertexSet|$. Each node $v \in \vertexSet$ has costs $\unaryCosts \in \mathbb{R}$ reflecting the likelihood of $v$ being a correct detection.  An edge $e=\{u,v\} \in \edgeSet$ encodes a possible linking of two detections to the same person. The nodes $u,v$ are labeled $\arbLabel$, if $\vertu$ and $\vertex$ belong to person $\arbLabel$. Likewise, $\pairwCosts \in \mathbb{R}$ reflects how likely $u$ and $v$ belong to the same person. 

Finally, the goal of the data association problem is then to find the labeling for all detection nodes that minimizes the total costs. 

Hence, for each node $\vertex \in \vertexSet$, consider a decision variable $\arbdecisionVar$ that equals $1$, if node $\vertex$ has label $\arbLabel$, and $0$ otherwise.  For $\numPersons$ being an upper bound on the number of persons,  let $[\numPersons]:=\{1,\ldots,\numPersons\}$.  Then, the vector $\vect{x} \in [0,1]^{n\numPersons}$ stacks all  decision variables in a vector. %

Given the unary and pairwise potentials
\begin{equation}
\mathrm{un}_{\graph}(\vect{x}) \coloneqq \sum_{v \in \vertexSet, \arbLabel \in [\numPersons]}\unaryCosts{\arbdecisionVar}
\end{equation} 
and 
\begin{equation}
\mathrm{pa}_{\graph}(\vect{x}) \coloneqq \sum_{\{\vertu,\vertex\} \in \edgeSet, \arbLabel \in [\numPersons]}{\pairwCosts \decisionVar_{\vertu}^{\arbLabel}\arbdecisionVar},
\end{equation} we define the cost function 
\begin{equation}
f_{\graph}(\vect{x})=\mathrm{un}_{\graph}(\vect{x})+\mathrm{pa}_{\graph}(\vect{x}).
\end{equation} 
Finally, our tracking model BQP$(\graph,\numPersons)$ is described by the labeling problem:
\begin{equation}
\label{eqn:discOpt}
\text{BQP}(\graph,\numPersons)\coloneqq \argmin_{\vect{x} \in \mathcal{C}_{b}(\graph,\numPersons)} f_{\graph}(\vect{x}), 
\end{equation} 
where $\mathcal{C}_{b}(\graph,\numPersons)\coloneqq \{0,1\}^{n\numPersons} \cap \mathcal{C}(\graph,\numPersons)$ and 
\begin{equation}
\label{eqn:constrSet}
\mathcal{C}(\graph,\numPersons)\coloneqq \{\vect{x} \in [0,1]^{n\numPersons} | \sum_{\arbLabel \in [\numPersons]}{\arbdecisionVar } \leq 1,\;  \forall \vertex \in \vertexSet \}.
\end{equation}
The constraints \eqref{eqn:constrSet} ensure that each detection is assigned to at most one label $\arbLabel \in [\numPersons]$, \ie to at most one person. 
Note that BQP$(\graph,\numPersons)$ has only $n$ linear constraints. 

We model the binomial distribution for the selection of nodes $\vertex$  and edges $e$ using logistic regression. Then, finding the most likely selection is equivalent to solving BQP$(\graph,\numPersons)$, if the costs are defined via the $\logit$ function, see \cite{tang2015subgraph}.  
Therefore, we set the unary costs as 
\begin{equation}
c_v\coloneqq\log\left((1-p_{\vertex})p_{\vertex}^{-1})\right)
\end{equation}
with $p_{v}$ denoting the probability of detection $v$ (inferred from the detection's score). For the pairwise costs,
 we learn model parameters $\theta$ to obtain probabilities
 \begin{equation}
 p_{\vertu,\vertex} \coloneqq  p(\decisionVar_{\vertu}^{\arbLabel}= 1 \land \arbdecisionVar= 1 |\modelFeatures,\theta),
 \end{equation}
given
 $\theta$ and a feature vector $\modelFeatures$. Since we model $ p_{\vertu,\vertex}$ using logistic regression, the pairwise costs are
 \begin{equation}
 \label{eqn:affMapping}
 \pairwCosts  \coloneqq \log((1-p_{\vertu,\vertex})p_{\vertu,\vertex}^{-1}).
 \end{equation}
 In Sect.~\ref{sec:features}, we
 describe the model features $\modelFeatures$ used for the classifier.

Detections, which are temporally too far apart can neither be compared reliably nor meaningful. For such edges $\{\vertu,\vertex \}$, we set their weight to $\pairwCosts  \coloneqq 0$. %
This strategy effectively sparsifies the graph $\graph$ and keeps the proposed approach memory and computationally efficient. 

\subsection{Frank-Wolfe Optimization} 
Solving BQP$(\graph,\numPersons)$ is a challenging task due to the fact that it
belongs to the NP-hard problems \cite{sahnisiam1974} and that our domain space is very high-dimensional.
We thus follow a common practice and consider the relaxed problem: 
\begin{equation}
\label{eqn:discOptProblem}
\text{QP}(\graph,\numPersons)\coloneqq \argmin_{\vect{x} \in \mathcal{C}(\graph,\numPersons)} f_{\graph}(\vect{x}) .
\end{equation}
 However, even the relaxation
is still NP-hard to solve \cite{pardalos1991quadratic}, as $f_{\graph}$ is non-convex, in general. Thus even for commercial quadratic solvers like Gurobi \cite{Gurobi}, solving BQP$(\graph,\numPersons)$ or QP$(\graph,\numPersons)$  is computationally very expensive.

This paper proposes to use the Frank-Wolfe algorithm to approximate QP$(\graph,\numPersons)$, and points out ways to further improve the solution.  We present a pseudo-code of the standard Frank-Wolfe
algorithm, together with a discretization step, in Alg.~\ref{alg:FW} and its evaluation, as a baseline,  in Sect.~\ref{sec:experiments}.

\begin{algorithm}
	\small
\SetAlgoLined
 \KwData{Costs $f_{\graph}$, feasible point $\iterateVecIndexed{0}$, $\IMAX$,$\epsilon$}
 \KwResult{Solution vector $\vect{x}_{\FW}$}
$f_{min} = \infty$\;
$j = -1 $\;
 \Repeat{$[(\atomicVecIndexed{j}-\iterateVecIndexed{j}^{\intercal}(-\nabla f_{\graph}(\iterateVecIndexed{j})) < \epsilon] \lor [j
   > \IMAX]$  }{
$j = j +1$ \;
$\atomicVecIndexed{j}= \argmin_{\atomicVec \in \mathcal{C}(\graph,\numPersons)} \atomicVec^{\intercal} \nabla f_{\graph}\bigl(\iterateVecIndexed{j}\bigr) $ \;
$\gamma(j) = \argmin_{\gamma \in  [0,1]} f_{\graph}\left(\iterateVecIndexed{j} + \gamma (\atomicVecIndexed{j}-\iterateVecIndexed{j})\right)$ \;
\If{$f_{\graph}(\atomicVecIndexed{j}) < f_{min}$}{$f_{\min} = f_{\graph}(\mathbf{s}(j))$\; $\vect{x}_{\FW} =
  \atomicVecIndexed{j}$ \;}
$\iterateVecBinarizedIndexed{j}= \round(\iterateVecIndexed{j})$ \;
\If{$f_{\graph}(\iterateVecBinarizedIndexed{j}) < f_{min}$}{$f_{\min} = f_{\graph}(\iterateVecBinarizedIndexed{j})$ \;
$\mathbf{x}_{\FW} = \iterateVecBinarizedIndexed{j}$ \;}
$\iterateVecIndexed{j+1}= \iterateVecIndexed{j}+\gamma(j)(\atomicVecIndexed{j}-\iterateVecIndexed{j})$\;
} 
\caption{Frank-Wolfe Algorithm}
\label{alg:FW}
\end{algorithm}

 \begin{figure*}[t!]      
\centering      
\def\svgwidth{0.835\linewidth} 
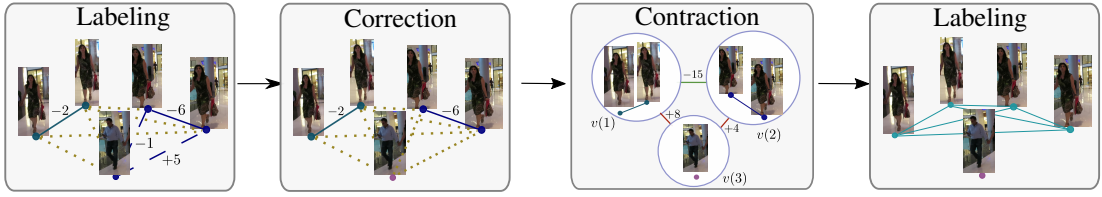
\caption{One iteration of our hierarchical solver. We present an illustrative example with pairwise costs on the edges. \textbf{Left}: A labeling, computed by Alg.~1. Dotted edges indicate removed links. Nodes connected by edges of the same color are grouped, resulting in two clusters.  Dashed edges indicate wrong connections.  \textbf{Middle}: Wrongly connected nodes (the men's node has positive costs of $+4$ to its connected nodes) are separated.  Then, each cluster is replaced by a new node (blue circles). The green edge indicates the correct assignment, whereas red edges indicate that clusters do not belong together. \textbf{Right}: The problem BQP$(\graph_{t+1},J)$ is solved w.r.t.\ $\graph_{t+1}$ and $J=\{1,2,3\}$ labels. Since $J$ is small, we can solve  BQP$(\graph_{t+1},J)$ quickly and optimal, using Gurobi. }
\label{fig:initFW}
\end{figure*}

Frank-Wolfe minimizes the linear approximation of $f_{\graph}$ at the current solution $\iterateVecIndexed{j}$ (Ln.~5 of Alg.~\ref{alg:FW}), resulting in $\atomicVecIndexed{j}$. The next iterate $\iterateVecIndexed{j+1}$ is the vector between $\iterateVecIndexed{j}$ and
$\atomicVecIndexed{j}$ that minimizes $f_{\graph}$ (Ln.~6 and Ln.~16).
For the optimal step size $\gamma(j)$ in Ln.~6, we present an efficient algebraic description in Sect.~\ref{subsec:stepSize}.  The algorithm is stopped in case of a small duality gap $(\atomicVecIndexed{j}-\iterateVecIndexed{j})^{\intercal}(-\nabla f_{\graph}(\iterateVecIndexed{j}))$ or a maximal number of iterations $\IMAX$. 

The binary solution $\mathbf{x}_{\FW}$ equals either a binarized iterate $\iterateVecIndexed{j}$ (Ln.~11-15), or, $\atomicVecIndexed{j}$ (Ln.~7-10),  
as the constraint matrix corresponding to our set $\mathcal{C}(\graph,\numPersons)$ is totally unimodular, so that  $\atomicVecIndexed{j}$ is already binary and thus feasible \cite{schrijver1998theory,joulin2014efficient}.  

In order to enhance the convergence rate, we use in our implementation  a slightly improved variant of the algorithm, that adds so-called away-steps. We
refer the interested reader to \cite{lacoste-julien_global_2015} for further details.

Using 
$\qMat_{\mathrm{pa}}=(\pairwCosts)_{\vertu,\vertex \in \vertexSet} \in \mathbb{R}^{n \times n}$ and $\mathbf{c}_{\mathrm{un}}= (\unaryCosts)_{\vertex \in \vertexSet} \in \mathbb{R}^n$, we define
\begin{equation}
\qMat=\text{diag}(\underbrace{\qMat_{\mathrm{pa}},\ldots,\qMat_{\mathrm{pa}}}_{\numPersons \text{ times}})
, \quad 
\vect{c} = (\underbrace{\vect{c}_{\mathrm{un}}^{\intercal} 
	,\ldots,\vect{c}_{\mathrm{un}}^{\intercal}}_{\numPersons \text{
		times} })^\intercal.
\end{equation}
\vspace{-0.01cm}Then, we obtain $f_{\graph}$ in matrix-vector form:
\begin{equation}
\label{eqn:fMat}
f_{\graph}(\vect{x}) = 0.5\vect{x}^{\intercal}\qMat\vect{x}+\vect{c}^{\intercal}\vect{x}.
\end{equation}
\vspace{-0.01cm}Due to design of our problem BQP$(\graph,\numPersons)$, we can run Alg.~\ref{alg:FW} without the need of storing the huge $\qMat$ matrix or the $\mathbf{c}$ vector.
Instead, all computations of Alg.\ref{alg:FW} can be deduced from the upper triangle matrix of $\qMat_{\mathrm{pa}}$ and from $\mathbf{c}_{\mathrm{un}}$. Therefore, our approach is memory efficient.

\noindent{\bf BINARIZE:}
\label{subsec:rounding}
In order to obtain  feasible, binary vectors, 
we discretize an iterate $\iterateVecIndexed{j}$ by selecting the closest feasible point $\iterateVecBinarizedIndexed{j}$ in $\mathcal{C}_{b}(\graph,\numPersons)$ \wrt euclidean distance. To this end, let $\mathbf{1}$ be the vector with all entries equal to $1$.  It is straightforward to show that 
\begin{align}
\label{eqn:roundingOpt}
\iterateVecBinarizedIndexed{j}&= \argmin_{\vect{x} \in \mathcal{C}_{b}(\graph,\numPersons) }{||\iterateVecIndexed{j}-\vect{x}||_{2}^{2} } \\
&= \argmin_{\vect{x} \in \mathcal{C}_{b}(\graph,\numPersons) }{(-2\iterateVecIndexed{j}+\mathbf{1})^{\intercal}\vect{x}}. \label{eqn:roundSquaredSimplified}
\end{align}

Now problem \eqref{eqn:roundSquaredSimplified} is linear in $\vect{x}$ and the constraint matrix corresponding to $\mathcal{C}(\graph,\numPersons)$ is totally unimodular. Thus,  we can efficiently solve the relaxation of \eqref{eqn:roundSquaredSimplified} and obtain the exact solution of \eqref{eqn:roundingOpt}, see, e.g. \cite[Chapter~19]{schrijver1998theory}.

\subsection{Computing the Optimal Step Size $\gamma$}
\label{subsec:stepSize}
The step size $\gamma$ in Ln.~6 of Alg.\ref{alg:FW} can be computed via line
search \cite{bertsekas1999nonlinear}. However, we derive a new algebraic computation, being faster and still optimal.

Let $\diffVec\coloneqq \atomicVecIndexed{j}-\iterateVecIndexed{j}$ and  
$\Omega(\gamma) \coloneqq f_{\graph}(\iterateVecIndexed{j}+\gamma
\diffVec)$.
Then, since $f_{\graph}$ is a quadratic, the only root of $\Omega'$ is $\gamma_{*}$, with
 \begin{equation*}
\gamma_{*} \coloneqq [-\diffVec^{\intercal}\nabla
f_{\graph}(\iterateVecIndexed{j})][\diffVec^{\intercal}\qMat \diffVec]^{-1}
\end{equation*}  
and  $\delta  \coloneqq \Omega''(\gamma_{*}) = \diffVec^{\intercal}\qMat \diffVec.$
Now if $\delta \neq 0$, the minimum $\gamma(j) = \argmin_{\gamma \in [0,1]}{\Omega(\gamma)}$ is given by
\small
\[\gamma(j) =\begin{cases}
\gamma_{*}, &\text{if } \delta > 0 \text{ and } \gamma_{*} \in [0,1], \\  
0,  & \text{if }(\delta > 0 \text{ and } \gamma_{*} \leq 0) \text{ or } (\delta < 0 \text{ and } \gamma_{*} \geq 1), \\
1, & \text{if }(\delta > 0  \text{ and  } \gamma_{*} \geq 1) \text{ or } (\delta < 0  \text{ and  } \gamma_{*} \leq 0), \\
\underset{\gamma \in \{0,1\}}{\argmin}\;\Omega(\gamma),  & \text{if } \delta < 0  \text{ and  } \gamma_{*} \in (0,1).
\end{cases}\]
\normalsize 
A line search is needed only if $\delta = 0$, making the execution of Ln.~6 very
efficient. In contrast to previous works \cite{DBLP:journals/corr/abs-1207-4747,dehghan_binary_2016},
our solution to Ln.~6 contains all cases that may occur.

\subsection{Regularization of the Objective Function}             
Since our cost function is non-convex, Frank-Wolfe delivers only a local optimum \cite{lacoste2016convergence}.  
Given $\regulizer \neq 0$, our next proposed improvement is to replace the objective function $f_{\graph}$ by 
\begin{equation}
f_{\regulizer}(\vect{x}) =f_{\graph}(\vect{x})+\regulizer\sum_{i}{(\decisionVar_i^2-\decisionVar_i)}.
\end{equation}
For $\vect{x} \in \{0,1\}^{n\numPersons}$, we have $f_{\regulizer}(\vect{x}) = f_{\graph}(\vect{x})$. Using $\regulizer <0$ has the effect of pushing the FW algorithm towards discrete solutions, as $-(\decisionVar_i^2-\decisionVar_i)$ has its minimum at $0$ and $1$, within $[0,1]$. For $\regulizer>0$, we observed better behavior in staying out of local optima, as for a value $\regulizer$  sufficiently large, $f_{\regulizer}$ becomes convex \cite{billionnet2012extending,burer2012non,hammer1970some}. On the other hand, a high $\regulizer$ value brings the optimal solution too close to the constant $(\frac{1}{2})$ vector.
For  $\omega:=\max\{{\sum_{j}|\qMat_{i,j}|}\, : i\in [n\numPersons]\}$, we set
 $\regulizer_{0}= \sqrt{\omega}$ and $\regulizer_{i} = 2^{-i} \regulizer_{0}$. Starting with
 $\regulizer=\regulizer_{0}$, we compute QP$(\graph,\numPersons)$,
 using $f_{\regulizer}$ in Alg.~\ref{alg:FW}. Empirically, we observed that a short number of iterations of Alg.~\ref{alg:FW} corresponds to a too strong convexification term, resulting in a bad local optimum. Thus, if Alg.~\ref{alg:FW} terminates in
 too few steps (which we set to 10), we set $i = i+1$, $\regulizer = \regulizer_{i}$ and run Alg.~\ref{alg:FW} again with the updated function $f_{\regulizer}$. In all our experiments, an appropriate $\regulizer$ was found in at most two calls of Alg. \ref{alg:FW}.  
In Sect.~\ref{sec:experiments}, we demonstrate the impact of using the modified cost function, with the solver we call $\mathbf{FW+\regulizer}$.

\subsection{Hierarchical Solving Scheme}
Since $FW+\regulizer$ delivers only a local optimum, 
we propose a new hierarchical solving
scheme that enhances the solution of $FW+\regulizer$ by
removing, correcting and connecting clusters, thus resulting in an improved objective value. Our approach is computationally efficient and continues optimizing problem BQP$(\graph,\numPersons)$. 
Compared to other hierarchical approaches like \cite{huang2008robust} that define specific parameter changes in each iteration,  our formulation is generic and can be applied to many clustering problems without the need of heuristically set parameter update rules.

In the following, we present all parts of our proposed solving scheme and present a pseudo-code in  Alg.~\ref{alg:hierarc}.

\noindent{\bf CorrectionContraction:} Let $\hierarchicalIterate{t} \in \mathbb{R}^{n\numPersons}$ be the current best labeling of $\graph$. Initially, we obtain $\hierarchicalIterate{0}$ using $FW+\regulizer$. 
We apply a relabeling strategy that corrects obvious errors within the clusters that may have been introduced due to the rounding or local optimality. For $\vertex \in \vertexSet$, let  $\neighborhood(\vertex,\hierarchicalIterate{t})$ be the set of all adjacent nodes that have the same label as $v$. If $\sum_{\vertu \in \neighborhood(\vertex,\hierarchicalIterate{t})}{\pairwCosts}>0$, or, if $\unaryCosts>0 $ and $(\hierarchicalIterate{t})_{\vertex}^{j} = 0,  \forall j \in [P]$, we 
assign a new and unique label to $v$ (see Fig.\ref{fig:initFW} middle). %
Let $\vertex(\arbLabel)$  be the set comprising all nodes labeled $\arbLabel$. We build a \textit{contracted graph} $\graph_{t+1}=(\vertexSet_{t+1},\edgeSet_{t+1})$ by using these virtual, new nodes: We set  $\vertexSet_{t+1}:=\{\vertex(\arbLabel)\mid \arbLabel \in [\numPersons]\}$ and $\edgeSet_{t+1}$ connects any two different vertices. 
Accordingly, we obtain the stacked decisions variables $\vect{x}_{\mathrm{contr}}$ for the current labeling of $\graph_{t+1}$. 

\noindent{\bf LabelExpand:} Let the current labeling result in $J$ clusters. To compute the optimal labeling on $\graph_{t+1}$ according to BQP$(\graph_{t+1},J)$, we define the unary costs
\begin{equation}
\label{eqn:unaryContr}
\underline{c}_{\vertex(\arbLabel)} \coloneqq  \sum_{\vertex \in \vertex(\arbLabel)}
{\unaryCosts}+\sum_{\{\vertu,\vertex\} \in \edgeSet \cap \vertex(\arbLabel) \times \vertex(\arbLabel)}{\pairwCosts}
\end{equation}
and pairwise costs 
\begin{equation}
q_{\vertex(k),\vertex(k')} \coloneqq \sum_{\{\vertu,\vertex \}\in \edgeSet \cap \vertex(\arbLabel) \times \vertex(k') }{\pairwCosts}.
\end{equation}
Consider the stacked decision variables $\hat{\vect{x}} \in \{0,1\}^{JJ}$ where $\hat{x}^{s}_{\vertex(\arbLabel)}$ equals 1, if $s = \arbLabel$ (and thus $(\vect{x}_{\mathrm{contr}})_{\vertex(\arbLabel)}^{s}=1$) and if $\vertex(\arbLabel)$ was not a rejected node by $\hierarchicalIterate{t}$; and $0$ otherwise.  Then, $\hat{\vect{x}}$ assigns each node of $\graph_{t+1}$ a unique label, except for nodes that have been rejected by $\hierarchicalIterate{t}$.
Therefore, $f_{\graph_{t+1}}(\hat{\vect{x}})$ sums up only the unary costs \eqref{eqn:unaryContr}, 
which equal $f_{\graph}(\hierarchicalIterate{t})$ or are improved by the refinement, implying
\begin{equation}
f_{\graph_{t+1}}(\hat{\vect{x}}) \leq f_{\graph}(\hierarchicalIterate{t}).
\end{equation}
Furthermore, solving BQP$(\graph_{t+1},J)$ results in a solution
  $\vect{x}_{\mathcal{H}} \in~\mathcal{C}_{b}(\graph_{t+1},J)$ with 
\begin{equation}
f_{\graph_{t+1}}(\vect{x}_{\mathcal{H}}) \leq f_{\graph_{t+1}}(\hat{\vect{x}}) \leq f_{\graph}(\hierarchicalIterate{t}).
\end{equation}   
The result $\vect{x}_{\mathcal{H}}$ is converted to a labeling $\hierarchicalIterate{t+1} \in \mathcal{C}_{b}(\graph,\numPersons)$ by \textit{graph expansion}: All nodes $\vertex \in \vertex(\arbLabel)$ are assigned the new label of $\vertex(\arbLabel)$, according to $\vect{x}_{\mathcal{H}}$, see also Fig.~\ref{fig:initFW}.
Thus, the hierarchical step can improve the last solution, since 
\begin{equation}
f_{\graph}(\hierarchicalIterate{t+1}) = f_{\graph_{t+1}}(\vect{x}_{\mathcal{H}})  \leq f_{\graph}(\hierarchicalIterate{t}).
\end{equation}

The graph contraction reduces the dimensionality significantly: There are $J \ll n $ nodes to be labeled using at most $J$ labels, w.r.t.\ BQP$(\graph_{t+1},J)$. If $J$ is small enough, we can solve
BQP$(\graph_{t+1},J)$ quickly to optimality using Gurobi 
\cite{Gurobi}. Otherwise we use the $FW+\regulizer$ solver.
The algorithm is stopped once no new clusters are merged.
 We demonstrate the effect of the hierarchical solving scheme $\mathbf{FW+\regulizer+h}$ in
 Sect. \ref{sec:experiments}. 
      \begin{algorithm}[t]
 	\small
 	\KwData{Graph labeling $\hierarchicalIterate{0}$, graph $\graph$}
 	\KwResult{Solution vector $\hierarchicalIterate{t}$}
 	\Repeat{$f_{\graph}(\hierarchicalIterate{t}) =  f_{\min}$}{	
 		$f_{\min} = f_{\graph}(\hierarchicalIterate{t})$\;
 		$(\graph_{t+1},\mathbf{x}_{\mathrm{contr}})=\text{CorrectionContraction}(\hierarchicalIterate{t},\graph)$\;
 		$\hierarchicalIterate{t+1}=\text{LabelExpand}(\graph_{t+1},\vect{x}_{\mathrm{contr}})$\;
 		$t = t+1$\;
 	}
 	\caption{Hierarchical solving scheme}
 	\label{alg:hierarc}
 \end{algorithm}


\section{Regression Training}
\label{sec:features}
In the following, we introduce spatial and temporal costs, which describe how likely two detections within the same and between different frames belong to the same person, respectively. For each cost type, we train a logistic regression model to obtain weights $\theta$, as described in Sect.~\ref{subsec:JDA}.

For our tracking system, we consider two input sources: (i)~head and (ii) full-body detections (see also Fig.~\ref{fig:teaserReal}).

\noindent{\bf Head detections.} To obtain accurate head detections, we employ
\cite{stewart_end--end_2015} based on Convolutional Neural Networks and
fine-tune it on the MOT16 training set \cite{MOT16}. 

\noindent{\bf Full-body detections.} We use the full-body detections \cite{felzenszwalb2010object} as provided by
the MOT16 challenge \cite{MOT16}.

\noindent{\bf Relative positioning.} In order to obtain meaningful features between differently sized boxes, features have to be formulated respecting the different  scales. %

\begin{figure}[t!]
\vspace{-0.1cm}
\centering  
\def\svgwidth{0.98\linewidth}
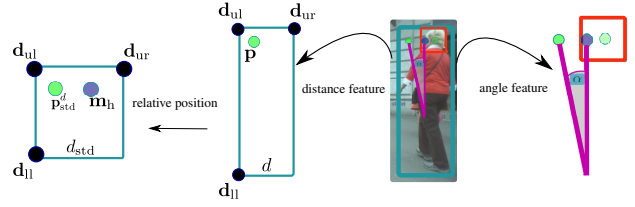
\caption{Distance and angle between the expected 
  (blue node) and the observed (mirrored) head position (green node). Barycentric coordinates are computed w.r.t.\ black corners.} 
\label{fig:angle}  
\end{figure}

To this end, consider a person detection box $d$ with the positions of lower left, upper left and upper right corners $\vect{d_{ll}},\vect{d_{ul}}$ and $\vect{d_{ur}}$, respectively and  $\Delta(d):=\left(\vect{d_{\mathrm{ll}}},\vect{d}_{\mathrm{ul}},\vect{d}_{\mathrm{ur}}
\right)^{\intercal}$.
For a pixel $\vect{p} \in \mathbb{R}^2$, we obtain barycentric coordinates $\vect{\lambda}_{d}=(\lambda_{1},\lambda_{2},\lambda_{3})^{\intercal}$ of $\vect{p}$ \wrt~$\Delta(d)$, so that $\vect{p} = \vect{\lambda}_{d}^{\intercal} \Delta(d)$ (see Fig.~\ref{fig:angle}).  We fix a standard box $d_{\mathrm{std}}$. 
Then, $\vect{p}$ is mapped to $\vect{p}_{\mathrm{std}}^{d} \coloneqq \lambda_{d}^{\intercal}\Delta(d_{\mathrm{std}})$, keeping the relative position as in $d$. Now, all subsequent distance measurements are computed using the mapped position \wrt $d_{\mathrm{std}}$.

\noindent{\bf Spatial costs.} 
We introduce two features that set the position of the head in relation to 
the full-body box. 
For a pair of head and full-body detection, we mirror the head detection to the left half side of the detection box $d$, resulting in the pixel $\vect{p}$, thereby making the position robust against different orientations of the person. From the MOT16
training data, we learned the expected relative position $\vect{m}_{\mathrm{h}}$ of a head \wrt the standard detection $d_{\mathrm{std}}$, corresponding to a full-body detection of the same person. Finally, we obtain the feature $\norm{\vect{p}_{\mathrm{std}}^{d}-\vect{m}_{\mathrm{h}}}$, measuring distance between the detected and expected
position. %
We introduce a second feature which uses the angle between expected and detected position, with
the anchor at the box's center (see \Fig~\ref{fig:angle}). 

We set the spatial costs between detections from the same detector to a constant high value.

\noindent{\bf Temporal costs.} 
Temporal costs are defined via correspondences of pixels between two frames. DeepMatching \cite{weinzaepfel:hal-00873592} (DM) provides such assignments, which are more reliable than spatio-temporal
affinities, see \cite{tangBMTT2016}. Given rectangles $u$ and $v$, DM samples $\mathrm{dm}_{u}$ and $\mathrm{dm}_{v}$ many pixels in $u$ and $v$, respectively.  Let $\text{co}_{u,v}$ denote the number of correspondences, found by DM. 
Comparing two heads or two full-body detections, we use the features $\frac{\mathrm{co}_{u,v}}{\mathrm{dm}_{u}}$,$\frac{\mathrm{co}_{u,v}}{\mathrm{dm}_{v}}$ and $\frac{\mathrm{co}_{u,v}}{0.5(\mathrm{dm}_{u}+\mathrm{dm}_{v})}$, as in \cite{tangBMTT2016}. As head detections are significantly  smaller than full-body detections, we only use the temporal head to full-body feature  $\frac{\mathrm{co}_{u,v}}{\mathrm{dm}_{u}}$, where $u$ denotes the head detection and $v$ the full-body detection. From the MOT16 training data, we learned the mean ratios $\phi^{w}_{\text{m}}$ and $\phi^{h}_{\text{m}}$ between a head and body detection, w.r.t.\ width and height, respectively, if both belong to the same person.  Then, we obtain features $\norm{\phi^{w}_{\text{m}}-\phi^{w}_{\text{det}}}$ and $\norm{\phi^{h}_{\text{m}}-\phi^{h}_{\text{det}} }$, for the observed ratios $\phi^{w}_{\text{det}}$ and $\phi^{h}_{\text{det}}$ w.r.t.\  width and height, respectively,  given a pair of detected head and full-body detection.


\section{Experimental Results}
\label{sec:experiments}

In this section, we first analyze the gain both in speed as well as in tracking performance by our proposed solver. 
Next, we investigate the impact of the detector fusion on the tracking performance, using the training sequences of the challenging MOT16 benchmark \cite{MOT16}. This benchmark consists of 7 sequences for training and 7 for testing, with footage of crowded scenes. 
In the last experiment, we show our performance on the test set of the benchmarks \MOTNEW and \MOTLATEST, where we achieve state-of-the-art performance. We evaluate our experiments using well-established tracking metrics \cite{bernardin2008evaluating,li2009learning,ristani2016performance}.
\subsection{Implementation Details}
\label{sec:implDetails}

In our implementation, we set the temporal costs of two nodes being more than $9$ frames apart to zero. The maximal number of labels $\numPersons$ is fixed to $70$.  We process a sequence in batches containing no more than $1800$ nodes.
We stop the Frank-Wolfe iterations of Alg.\ref{alg:FW} in case the duality gap is below $10^{-4}$ or $750$ iterations are reached.

\subsection{Frank-Wolfe Optimization}
\label{FWexperiments}

Our first experiment analyzes the 
 impact of our modifications on the Frank-Wolfe optimization.
To this end, we choose a representative batch of $41$ frames from the MOT16-13 training sequence and perform tracking using full-body detections only. It consists of $403$ detections, so that we have  $28210$ decision variables. 
In Tab.~\ref{fig:BQPTable} we show the number of iterations performed by the solver until the duality gap is below the defined threshold, the runtime, the final objective value of $f_{G}$ as well as the corresponding Multiple Object Tracking Accuracy (MOTA).

\begin{table}
\caption{Solver comparison: While our solver quickly terminates, Gurobi is not able to finish after $1000$ seconds. Entries in brackets denote the results of Gurobi after that time.}
\smallskip
\renewcommand{\arraystretch}{0.7}
\resizebox{1\linewidth}{!} {
\resizebox{1\linewidth}{!}{\begin{tabular}{ l c c c c }
\toprule
Method & Iters & Time[sec] & Obj Value & MOTA \\
\midrule
$FW$ & \textbf{16} & \textbf{0.7} & -3060 & 14.2 \\
$FW+\regulizer$ & 676 & 27 & -5481 & 26.8 \\
$FW+\regulizer+h$ & - & 27+0.5 & \textbf{-5925} & \textbf{27.5} \\
Gurobi & - & 1000 & (-5531) & 24.9 \\
Gurobi bound & - & 1000 & (-5973) & - \\
\bottomrule
\end{tabular}}}

\label{fig:BQPTable} 
\end{table}

Our proposed modification $FW+\regulizer+h$ improves the objective value considerably compared to the standard Frank-Wolfe algorithm $FW$. This naturally translates to almost double MOTA accuracy, 14.2\% vs 27.5\%. 
Note also that the objective value comes very close to the global optimum. The commercial solver Gurobi \cite{Gurobi}, which uses the branch-and-bound algorithm is still far away from the global optimum after 1000 seconds, while we obtain a much better energy after only $27.5$ seconds. While Gurobi was not able to compute the global optimum in the given time span,
it delivers at each time step a lower bound (Gurobi bound) on the optimal value, showing that the optimal solution to the BQP has an objective value $\geq -5973$. 

\begin{figure} 
	\centering
	\includegraphics[scale=0.35]{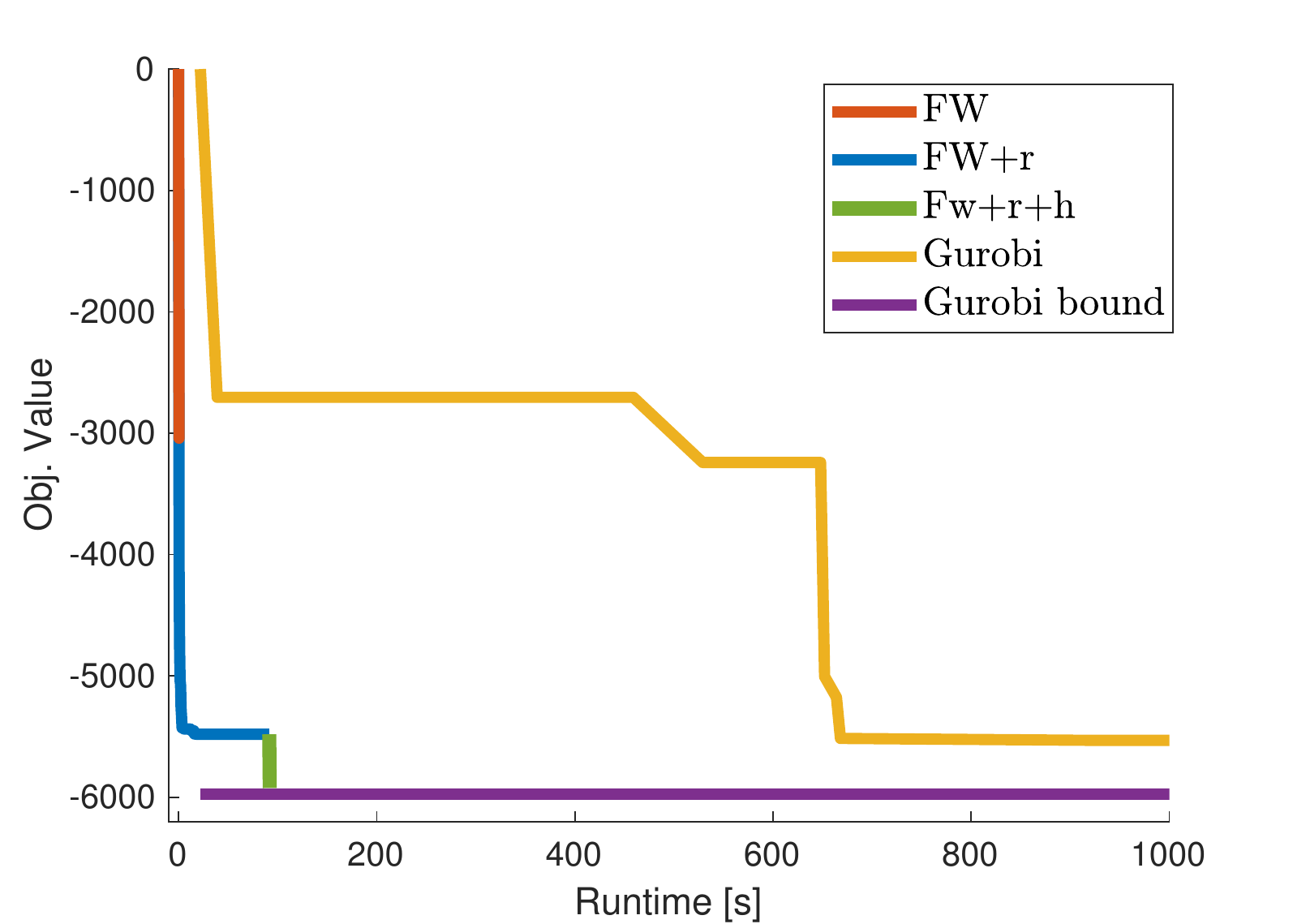}
	\caption{Minimization performance of each BQP solver and the bound as given by Gurobi for each moment.}
	\label{fig:BQPSolvers}  
\end{figure}

The energy evolution of the different solvers is plotted in Fig.~\ref{fig:BQPSolvers}. Here we clearly see where $FW$ stops (red line), how our modification $FW+\regulizer$ improves the energy by a large margin (blue line), and how finally $FW+\regulizer+h$ (green line) comes even closer to the estimated lower bound (purple line), as provided by Gurobi. In contrast, Gurobi (yellow line) has a much slower convergence.

To separate the quality of our solver from the detections, we further evaluate the performance on ground-truth person detections for 40 frames of each MOT16 training sequence in Tab.\ref{tab:USE_GT_COSTS_0_USE_GT=1_FWMODUS=1}, where we also report the (relative) duality gap to the optimal solution (GAP). The results show a consistent and huge improvement by the hierarchical concept over FW+\regulizer. At the same time, the solutions are close to optimality \wrt to the objective value and \wrt to tracking performance. The sequences MOT16-05 and MOT16-11 both contain many partial occlusions that makes it difficult for the DM features to be correct in any situation, thus resulting in lower tracking scores. However this shows that a second type of detections (head detections) is necessary for high quality tracking results.  On the other hand, the solver reaches the perfect result on MOT16-09 (which has far less occlusions), thereby justifying our solver.

\begin{table}[htp]
	\begin{center}
	\label{tab:USE_GT_COSTS_0_USE_GT=1_FWMODUS=1}
\tabcolsep=0.05cm
     \renewcommand{\arraystretch}{0.55}
     	\caption{FW+\regulizer\xspace versus FW+\regulizer+h (on GT detections).}
\resizebox{1\linewidth}{!}{\begin{tabu}{r|[1.5pt]c|c|c|c|c|[1.5pt]c|c|c|c|c}
		\toprule
		{} & \multicolumn{5}{c|[1.5pt]}{FW+\regulizer} & \multicolumn{5}{c}{FW+\regulizer+h}  \\
			{Seq} & {IDF1} & {ID} & {FM} & {MOTA} & {GAP} & {IDF1} & {ID} & {FM} & {MOTA} & {GAP}\\
			\midrule
							{02} & 87.4 & 5 & 1 & 84.0  &   6.424 & 90.9 & 3 & 0 & 90.8  &    0.428 \\
							{04} & 85.0 & 5 & 0 & 73.2 &  7.506 &   92.4 & 0 & 0 & 85.8 &     0.120 \\
							{05} & 57.4 & 10 & 8 & 74.2 &  9.130  &70.1 & 8 & 7 & 75.1  &     0.071 \\
							{09} & 80.6 & 3 & 0 & 98.9  &  5.353  & 100.0 & 0 & 0 & 100.0 &   0.000 \\
							{10} & 82.0 & 10 & 6 & 80.4 &  7.410  & 87.0 & 7 & 6 & 89.4  &    0.638 \\
							{11} & 76.8 & 13 & 2 & 78.2 &  12.846 & 89.4 & 5 & 3 & 96.3  &    0.084  \\
							{13} & 87.2 & 10 & 2 & 85.3 &  10.332 & 96.3 & 2 & 3 & 96.9  &    0.434  \\
		\bottomrule
		\end{tabu}}
	\end{center}
	\vspace{-0.3cm}
\end{table}

\vspace{-0.15cm}

\subsection{Ablation studies on head and body detections}
\label{ablation}
We analyze how our formulation exploits the information from two detectors. 
For this experiment, we use all MOT16 training sequences with the full-body
detections only (\textit{B}) against body and head detections
(\textit{B+H}). We use the body detections provided by the benchmark while we train the head detector and the regression model on MOT16 training sequences in a leave-one-out fashion. 

In \Tab~\ref{tab:trainSet}, we report several ablation results with: (i) different inputs (body and heads) and (ii)  different solvers, namely  our proposed FW+\regulizer+h (Ours) is compared to: tracking heads and bodies independently and then using our solver to fuse them. The tracklets are computed by our system (Ours-fusion) and from LP2D \cite{lealiccv2011} (LP2D-fusion). We use the affinities as defined in Sect.3, but set the spatial and temporal costs between two tracklets that originate from the same detector to a constant high value, as the tracklets are already separating the persons (Sect.3*).
We further provide the quality of the head trajectories, which we evaluated on the head ground truth boxes.

\begin{table}[ht]
\caption{Ablation experiments on MOT16 training sequences.}
\renewcommand{\arraystretch}{0.8}
\smallskip
\centering
\resizebox{1\linewidth}{!} {
 \begin{tabular}{@{}l c c c c c c c c @{}} 
\toprule
Feature & Affinities & Solver &SolverID & MOTA &  MT  & FP & FN & IDs    \\ 
\toprule
\toprule

H & 2D dist & $LP2D$& 1& 14.9   &{\bf 70} & 14829&  50991 &  472 \\
H & Sect.3 & $Ours$& 2& {\bf 16}   &{\bf 70} & {\bf 14168}&   {\bf 50959}& {\bf 331} \\

\toprule
B & 2D dist & $LP2D$&  3&31.7   &44 & {\bf 3557}&  71332&  467 \\
B & Sect.3 & $Ours$ & 4 &33.0   & {\bf 76} & 11949& {\bf 61603} & {\bf 378} \\
B & \cite{dehghancvpr2015} & GMMCP\cite{dehghancvpr2015}& 5 & {\bf 33.7} & 46  &   4053 &   68675  & 499 \\

\toprule
B+H & Sect.3*& $LP2D-fusion$&  6 & 33.0 &  54  & {\bf 3501} &    70163 & { 358}\\
B+H & Sect.3*& $Ours-fusion$&  7 &34.2 & {\bf 87}  &  11852 & {\bf60401}  & 376 \\
B+H & Sect.3 & $FW$&   8 &31.1 & 75  &  5315 &   69563  & 1207 \\
B+H & Sect.3 & $FW+\regulizer$&  9 & 33.4 & 82  &  6497 &   66238  & 807 \\
B+H & Sect.3 & $Ours$&  10 &{\bf 38.2} & 86  &  4972 &   62935  & 372 \\
B+H & Sect.3 & NLLMPa \cite{levinkov2017joint}&   11 &37.4 &  86  & 4954 &    63831  & {\bf 336} \\

\bottomrule 

\end{tabular} }
\label{tab:trainSet}
\end{table}

Our system performs comparable on full-body detections to the SolverID 5, using their defined affinities. 
By using the two detectors, our system significantly improves almost all
relevant tracking metrics, justifying our tracking framework (SolverID 10 vs 4). Due to the coupling of head detections with full-body detections, the number of false positives (FP) is halved and the system is less prone to partial occlusions, which results in an increase of the number of mostly tracked (MT) trajectories. Overall, the MOTA score
increases by more than 5pp (percentage points). Performing the fusion directly on the input detections is clearly more effective  than using initial tracklets.  SolverID 6 and 7 use our solver and precomputed trajectories from SolverID 3 and 4, where the gain is no more than 1.3pp, justifying our fusion concept. Using another heuristic solver \cite{levinkov2017joint} (SolverID 11) performed worse on the fusion than FW+\regulizer+h, using exactly the same graph. The comparison SolverID 8-10 show the improvement on MOT16train due the regularizer and the hierarchical step (up to 7.1 pp on the MOTA score).

\subsection{Benchmark Evaluation}
We evaluate the tracking performance of our formulation with body and heads on the benchmarks \MOTNEW and \MOTLATEST with the full-body detections provided by the benchmarks. Due to space constraints, we show some of the best performing published trackers in Tab.~\ref{tab:mot16}, as well as the worst performing tracker. For the full table of results, please visit the benchmark's website. 

Our system creates slightly higher identity switches. This can be resolved in future work with more advanced features that include a foreground/background mask in each detection or in a post-processing step where tracklet consistency is checked, though this is beyond the scope of this paper.
However, our proposed tracker performs on par with state-of-the-art in terms of tracking accuracy on \MOTNEW and sets a new state-of-the-art on \MOTLATEST.  Furthermore, the tracker won, together with \cite{keuper2016multi}, the MOT 2017 Tracking challenge at the CVPR 2017 \footnote{\url{https://motchallenge.net/MOT17_results_2017_07_26.html}}.
Note that the MOTA metric is regarded as the most representative metric \cite{leal2017tracking}. With our proposed formulation, we have the lowest ML (mostly lost) score within all trackers in both benchmarks, showing that we can recover more trajectories than any other tracker. Also our MT score is highest on the MOT16 benchmark and ranks second on the MOT17 benchmark, demonstrating that we recover very long trajectories.  
In contrast, the GMMCP model approach is not able to produce long-term consistent  trajectories possibly due to erroneous initial tracklets, that could not be connected (we used the official code of \cite{dehghancvpr2015} to produce the results). We note that the LMP tracker uses very advanced and stable convolutional neural network image features that can reliably link boxes over 200 image frames, thus resulting in a better MOTA score.

\begin{table}[b]
  \caption{Public tracking results on \MOTNEW and \MOTLATEST.}
  \label{tab:mot16}
  \smallskip
   \renewcommand{\arraystretch}{0.8}
\centering
\resizebox{1\linewidth}{!} {
\begin{tabular}{@{}l rrrrrrrrr @{}}
\toprule
  Method & Rank & MOTA$\uparrow$ & IDF1$\uparrow$ & MT$\uparrow$ & ML$\downarrow$ & FP$\downarrow$ & FN$\downarrow$ & ID$\downarrow$ \\ \midrule
&&&&  MOT16 &&&& \\ \midrule
LMP \cite{tang2017multiple} & 1 &{\bf 48.8}	&  51.3& 	18.2& 	40.1& 	6654	& 86245 & 	 481	\\
 {\bf Ours} &  2 & 47.8 & 47.8 & {\bf 19.1} & {\bf 38.2} &  8886 & {\bf 85487}  & 852  \\
NLLMPa \cite{levinkov2017joint} & 3 &47.6	& 47.3& 	17.0& 	40.4& 	 5844	& 89093 & 	629\\
AMIR \cite{Sadeghian:ICCV:2017} & 4 & 47.2&	46.3&	14.0&	41.6&	{\bf 2681}&	92856	&774 \\ 
NOMT \cite{choiiccv2015} & 5 & 46.4	&{\bf 53.3}&	18.3&	41.4&	9753&	87565&	{\bf 359} 	\\
GMMCP \cite{dehghancvpr2015} & 15 & 38.1	& 35.5&	8.6&	50.9&	6607&	105315&	 937 	\\
DP\_NMS \cite{pirsiavashcvpr2011} & 23 & 26.6 &	31.2	& 4.1&	67.5&	3689&	130557&	365 \\
\midrule
 &&&&  MOT17 &&&& \\ \midrule
 	{\bf Ours} & 1 & {\bf 51.3} &  47.6 & 21.4 & {\bf 35.2} &   24101 & 247921  & 2648  \\
 	MHT\_DAM\cite{Kim:2015:ICCV} & 2 & 50.7 & 47.2 & 20.8 & 36.9 &  {\bf 22875} & 252889  & 2314  \\
 	EDMT17\cite{ChenBMTT2017} &3  & 50.0 & {\bf51.3} & { \bf 21.6} &  36.3 &  32279  & {\bf 247297} & {\bf 2264}  \\
	 GMPHD\_KCF\cite{Kutschbach:AVSS:2017} &6  & 30.5 & 35.7 & 9.6 &  41.8 &  107802  & 277542 & 6774  \\
\bottomrule
\end{tabular}
}
\end{table}

\texttt{%
}

\vspace{-0.45cm}
 \section{Conclusion}
 \label{sec:conclusion}
 
We presented a global formulation for multi-detector multi-target tracking, and showed its state-of-the-art performance with head and full-body
detectors. We proposed to cast the problem into a quadratic
program, which is solved efficiently via the Frank-Wolfe algorithm. We
improved the solver in three ways; (i) regarding time by providing complete and
efficient computation of the optimal step size and (ii) regarding minimization by
a reformulation of the objective function, resulting in better discrete solutions.
Finally (iii), we showed that our hierarchical solving scheme improves a feasible
solution, often close to optimality and yet is easy to
integrate and fast.

The detector fusion delivered superior results when compared
to single detector tracking, thus proving the benefits of our formulation. The
overall performance on two challenging tracking benchmarks showed state-of-the-art results.

{\small
\bibliographystyle{ieee}
\bibliography{TrackFusion}
}

\end{document}